# Arap-Tweet: A Large Multi-Dialect Twitter Corpus for Gender, Age and Language Variety Identification

**Wajdi Zaghouani, [1] Anis Charfi [2]**

[1] College of Humanities and Social Sciences, Hamad Bin Khalifa University, Qatar
[2] Information Systems Program, Carnegie Mellon University Qatar
E-Mail: wzaghouani@hbku.edu.qa, acharfi@qatar.cmu.edu

## Abstract

In this paper, we present Arap-Tweet, which is a large-scale and multi-dialectal corpus of Tweets from 11 regions and 16 countries in the Arab world representing the major Arabic dialectal varieties. To build this corpus, we collected data from Twitter and we provided a team of experienced annotators with annotation guidelines that they used to annotate the corpus for age categories, gender, and dialectal variety. During the data collection effort, we based our search on distinctive keywords that are specific to the different Arabic dialects and we also validated the location using Twitter API. In this paper, we report on the corpus data collection and annotation efforts. We also present some issues that we encountered during these phases. Then, we present the results of the evaluation performed to ensure the consistency of the annotation. The provided corpus will enrich the limited set of available language resources for Arabic and will be an invaluable enabler for developing author profiling tools and NLP tools for Arabic.

## 1. Introduction

Arabic is a challenging language when it comes to building Natural Language Processing tools and applications. In fact, the complexity of Arabic is present at the various levels of linguistic representation (phonology, orthography, morphology, and syntax). Even though there were some advances in the field of Arabic Natural Language Processing, the Arabic language is still lagging behind other languages such as English in terms of availability of the required resources to address author profiling (Rosso et al., 2018; Zaghouani, 2014).

In fact, a person's language use reveals much about their profile. However, research on author profiling has always been constrained by the limited availability of training data, since collecting textual data with the appropriate meta-data requires a significant collection and annotation effort. For every text, the characteristics of the author have to be known in order to successfully profile the author. For the Arabic language, to the best of our knowledge, there is no corpus freely available for the detection of age, gender and dialectal variety. Most of the existing corpora are available for English or other European languages (Celli et al., 2013). Having a large amount of data remains the key to achieving reliable results in the task of author profiling.

This paper presents the work carried out within the framework of the Arabic Author Profiling Project (ARAP),[1] a research project funded by Qatar National Research Fund. This project aims at developing author profiling resources and tools for the Arabic language and using them in the context of cyber-security. More specifically, author profiling in the context of ARAP could be useful for forensic investigations to narrow the set of potential authors when receiving a threat message. While a few research efforts on author profiling have recently started in Europe and the USA, there is extremely little research that targets the Arabic language.

Within the context of the ARAP project, we built the Arap-Tweet corpus (Zaghouani & Charfi, 2018a), a multi-dialectal annotated corpus that can be used for author profiling, stylometry research, and many other applications. For instance, this kind of resources could be useful in studying Arabic dialects from a linguistics perspective (e.g., computational dialectology).

Twitter offers the opportunity to gather large amounts of informal language from many individuals. We searched and collected Twitter profiles from 11 regions in the Arab world in order to cover the most distinct dialectal varieties. Once the data collected, processed and normalized, we started the annotation process using well-defined annotation guidelines to annotate the Tweets according to their dialectal variety, the gender of the user and the age within three categories (under 25 years old, between 25 and 34, and above 35). Finally, we evaluated the quality of the data by performing the inter-annotator agreement measures on a regular basis during the whole process.

In the remainder of this paper, we briefly review related work (Section 2) and report on the dialectal Arabic varieties (Section 3). Then, we present our corpus and the respective data collection and validation processes (Section 4). After that, we present our annotation guidelines and workflow (Section 5). Finally, we present the evaluation of the annotation quality (Section 6).

## 2. Related Work

In the context of corpus creation for the modern standard Arabic, there are several efforts (Habash, 2010). In fact, there are many monolingual and parallel corpora annotated with syntactic and semantic information such as the different iterations of the Penn Arabic Probanks (Diab et al., 2008; Zaghouani et al., 2010; Zaghouani et al., 2012) and treebanks (Maamouri et al., 2010). Many tools and methods were developed to deal with the morphology, disambiguation (Zaghouani et al., 2016c), the diacritization (Zaghouani et al., 2016b) and syntactic parsing (Habash, 2010).

---

[1] http://arap.qatar.cmu.edu

For Dialectal Arabic (DA), some limited efforts were made to create resources for some major Arabic dialects such as Egyptian and Levantine (Habash et al., 2013; Diab & Habash, 2007; Pasha et al., 2014). Within the framework of the Qatar Arabic Language Bank (QALB) project, a large-scale annotated corpus of users' comments was produced, dialectal words were marked (Zaghouani et al., 2014; Zaghouani et al., 2015; Zaghouani et al., 2016a.)

Al-Sabbagh and Girju (2010) presented a method to extract information from the Internet in order to build a Dialectal to Modern Standard Arabic lexicon. Chiang et al. (2006) created a parser for Dialectal Arabic that was trained on MSA treebanks. Similarly, Sawaf (2010) worked on processing Dialectal Arabic using the training data from the standard Arabic Penn Treebank while Salama (2014) created an automatically annotated large-scale multi-dialectal Arabic corpus collected from user comments on Youtube videos. Their corpus covers five regions the Arab world, namely: Egypt, Gulf, Iraqi, Maghrebi and Levantine.

Some other works such as Sajjad et al. (2013), Salloum and Habash (2013) and Sawaf (2010) used a translation of the dialectal Arabic to Standard Arabic as a pivot to translate to English. Boujelbane et al. (2013) created a dictionary based on the relation between Tunisian Arabic and MSA. Some other researchers followed crowdsourcing based approaches to create interesting resources such as the work of Zbib et al. (2012).

At the regional level, we noted limited efforts focused on dialect identification such as in (Habash et al., 2008; Elfardy & Diab, 2013; Zaidan & Callison-Burch, 2013).

As the Dialectal Arabic (DA) is becoming the language of informal online communication in emails, chats, SMS and in social media, we witnessed several efforts on creating different resources to help to build related tools and applications. However, most of these efforts were disconnected from each other and they have only focused on a limited number of dialects in the Arab world.

To the best of our knowledge, there are only a few resources available on author profiling for the Arabic language and for the dialectal Arabic. We found two projects and two resources related to that topic: Abbasi and Chen (2005), Estival et al. (2008), Mubarak and Darwish (2014), and Rangel et al. (2017).

Abbasi and Chen (2005) focused on author identification in English and Arabic web forum messages in order to do an analysis of the extremist groups web forum messages. Estival et al. (2008) built the Text Attribution Tool (TAT) to automate the analysis of texts for the purpose of author profiling and identification for English and Arabic E-Mails. Mubarak and Darwish (Mubarak & Darwish, 2014) built a Twitter dialectal Arabic corpus from four different Arabic countries using the geolocation information associated with Twitter data. More recently, Bouamor et al. (2018) and Habash et al. (2018) created MADAR, an Arabic dialect corpus and lexicon covering dialects of various cities across the Arab world.

In the context of our ARAP project, co-organizers of the Author Profiling Share task PAN 2017[2] presented a dialectal Arabic corpus from four different regions (North Africa, Egypt, Levantine and Gulf). The corpus data were annotated with respect to age, gender and dialect (Rangel et al, 2017).

For both Mubarak and Darwish (2014) and Rangel et al. (2017), the coverage was limited to only four countries out of 22 Arabic countries. In the ARAP project, we extend this coverage to all Arabid major dialects by covering 11 distinct dialects. The provided corpus can be used for applications in various domains such as cyber-security, business (e.g., for marketing and customer segmentation) and in healthcare (e.g., for suicide prevention).

### 3. Dialectal Arabic

The Arabic language used in social media and online is a mix of Modern Standard Arabic (MSA) and other regional dialectal varieties. For this reason, it is important to recognize this code-switching situation when studying the Arabic language produced by users online.

Arabic dialects are generally classified by regions such as in Habash (2010) who classified the Arabic major dialects into North African, Levantine, Egyptian and Gulf or into sub-regional classification (e.g., Moroccan, Tunisian, Algerian, Egyptian, Lebanese, Syrian, Jordanian, Qatari, Iraqi etc.). Figure 1 illustrates the different dialectal varieties in the Arab world across the different countries and borders.

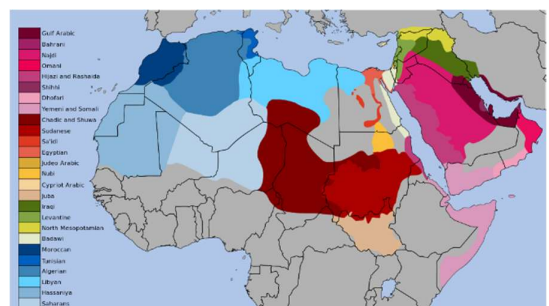

Figure 1: Arabic dialectal varieties in the Arab world[3]

The variation from one region to another poses many challenges to Natural Language Processing applications and therefore fine-grained resources and tools are required to address this issue. In fact, resources made for a given region cannot be used to train models for the dialect of a different region. While there are many similarities in the Arabic dialects, there is often a difference at the level of the lexicon, the morphology and the phonology.

---

[2] http://pan.webis.de/clef17/pan17-web/author-profiling.html

[3] Map distributed under a CC-BY 3.0 license from Wikipedia.

The sentence example in Table 1 illustrates similarities and differences between some dialectal varieties and standard Arabic. Within the ARAP project, we collected Tweets from 11 major Arabic regions or dialect groups instead of the traditional four groups in order to better represent the Arabic language used online in social media throughout the different regions.

| Variety | I love reading a lot. |
|---|---|
| Standard Arabic | أنا أحب القراءة كثيرًا<br>ʾanā ʾuḥibbu l-qirāʾata |
|  |  |
| Tunisian | āna nḥəbb năqra barʃa<br>أنا نحب نقرا برشة |
| Algerian | āna nḥəbb nəqra bəzzāf<br>أنا نحب نقرا بزاف |
| Moroccan | ana nbɣi bezzaf nəqra<br>أنا نبي بزاف نقرا |
| Egyptian | ana baḥebb el-ʔerāya awi<br>أنا بحب القراءة أوي |
| Lebanese | āna ktīr bḥebb il-ʔirēye<br>أنا كتير بحب القراءة |
| Iraqi | āni kolish aḥeb el-qra'a<br>أني كلش أحب القراء |
| Qatari | ʔāna kulliʃ aḥibb aqrā<br>أنا كلش أحب أقراء |

Table 1: A sample sentence in seven Arabic Dialects

## 4. Corpus Creation

In this section, we report on the corpus creation and annotation efforts we carried out to locate and crawl users for each dialect group. Before collecting and processing the data and in order to create the first Twitter multi-dialect annotated corpus of Arabic, we tried to cover as many dialects as possible while taking into consideration the available resources. We were able to collect Tweets from 11 Arabic regions representing a total of 16 countries from a total of 22 Arab countries that are members of the Arab league as shown in Table 2.

For each region, we collected the profiles of 100 users with at least 2000 Tweets with a minimum of 200K Tweets per region and a total of 2.4M Tweets corpus.

### 4.1 The Annotation Logistics

The collection and the annotation of a large scale corpus covering many regions of the Arab world such as the Arap-Tweet require the involvement of many annotators. In our project, the annotation effort is led by an annotation manager, and the team also consists of junior annotators and programmers.

The annotation manager is responsible for the whole annotation task. This includes collecting and cleaning the corpus, the annotation of a gold standard set to be used during the evaluation. Moreover, he is in charge of writing the annotation guidelines, hiring and training the annotators and monitoring the annotation progress and quality by performing the annotation evaluation on a weekly basis.

To ensure the suitability of the annotators for the task, we selected only university level annotators with a strong background knowledge of dialectal Arabic covering the regions to be annotated. Furthermore, the annotators were tested in a dialectal Arabic language screening test. Once selected, the annotators were trained over a period of two weeks by doing a pilot annotation task.

| Dialect | Region |
|---|---|
| Moroccan | Morocco |
| Algerian | Algeria |
| Tunisian | Tunisia |
| Libyan | Libya |
| Egyptian | Egypt |
| Sudanese | Sudan |
| Lebanese | North Levant |
| Syrian | North Levant |
| Jordanian | South Levant |
| Palestinian | South Levant |
| Iraqi | Iraq |
| Qatari | Gulf |
| Kuwaiti | Gulf |
| Emirati | Gulf |
| Saudi | Gulf |
| Yemeni | Yemen |

Table 2: Dialects and regions covered in the corpus

### 4.2 Selecting and Crawling Users

When looking at Twitter public profiles, we found that some users may include information such as their real name, location and a short biography in their profile. However, their age and gender details are usually not shared as they are not required in the Twitter profile and as there are no explicit fields on Twitter for them. Our goal was to select a balanced set of users for each of the 11 dialect regions selected. We found the users by searching public Tweets containing specific seed words and expressions that are used only in one dialect. For example, the word برشة /barsha/ 'many' in Tunisian Arabic or the word وايد /wayed/ 'many' in Gulf Arabic. In order to identify the list of seed words for each dialect, we conducted with the annotators a comprehensive study to validate the seed words list as some seed words were common in more than one region. Moreover, the annotators were trained to identify if a given seed word was used in the user profile from a different region. For instance, in order to accurately identify the profile for each region using the seed words, we relied on multiple seed words occurrences in multiple Tweets from the same profile in order to validate correct region for the given profile.

Once the potential users identified for a given dialect, the list was reviewed manually by the annotators to confirm and match the identified users to their dialect. We tried to select the data as randomly as possible by avoiding well-known public figures from our list. This restriction led to a large annotation effort and resulted in a smaller user sample.

Using the Twitter API we collected tweets that contained typical dialectal distinct words generally used by speakers of a given region. This allowed us to restrict the tweets to the selected region as much as possible. During a twelve weeks period, we sampled users according to this method.

We only included accounts with a minimum number of 2000 Tweets. For all users, we downloaded up to their last 3240 tweets (limit imposed by Twitter API). We excluded retweets from the data collection as these tweets were written by other people.

For the annotation, our annotators carefully analyzed the users' profiles and their tweets. They had to annotate the collected data with the age group, the gender, and the dialect. Whenever possible they also used external resources such as LinkedIn, Facebook, and the user's blog and web page. For profiles that had a photo the annotators used the photo to guess the age in addition to using the AI based Microsoft website How-Old.net.[4]

In order to produce a usable and clean corpus for the planned author profiling task, we documented our annotation guidelines (Zaghouani & Charfi 2018b) and we asked the annotators to only annotate users who meet the following requirements:

- The profile should belong to an actual person (e.g. not an association or a company).
- The profile should be publicly accessible.
- The profile should have at least 2000 tweets.
- The tweets should have been mostly written in the given dialect (from the list of the 11 dialects).
- The Tweets should not be written mostly in standard Arabic or any other language such as English or French (this requirement is validated manually by the annotator by going through the profile Tweets manually.)
- The profiles posting a lot of images and using applications to automatically post daily messages by bots were also filtered

During the profiles collection step, we noticed that a group of users decided to protect their account and make it private between the time of sampling and the time of data collection. In total, 1100 users were annotated (100 users per region).

### 4.3 Gender Annotation

The gender was annotated for 1100 persons. In some cases, the annotators were not able to identify the gender due to the lack of a profile photo or other identifying information. In such cases, the users were removed from our list and replaced by users of the same gender and from the same region.

The gender male/female ratio was almost equal for 7 regions while for 4 regions it was around 60% males and 40% females on average as for some Arab countries we noticed that Twitter was not widely used by females. The annotation of the gender was based in most cases on the name of the person or on his profile photo and in some cases on their biography or profile description.

### 4.4 Age Annotation

In order to annotate the users for their age, we used three categories: under 25 years, between 25 years and 34 years, and above 35 years. The age category was annotated for all 1100 accounts. The results separated by gender are shown in Table 3. There are more females in the young age group, while there are more men in the older age groups.

| Age Group | Male | Female |
|---|---|---|
| Under 25 | 150 | 94 |
| 25 until 34 | 391 | 199 |
| Above 35 | 126 | 140 |
| Total | 667 | 433 |

Table 3: Age/Gender Annotation Groups

For annotating the age, our annotators started by retrieving the real name of the twitter user if possible. If the user has a profile photo we asked the annotators to guess the age based on the photo and then to use the machine learning based Microsoft website How-Old.net. If there is no photo in the profile we asked them to check if the users have other web pages or social media accounts (e.g., on Facebook or LinkedIn). For LinkedIn users, the age group could be determined by checking the education history of the users if it is indicated. Also, for some Facebook users the age could be determined if they indicate the year at which they graduated from high school. If the annotators were not able to annotate the age we asked them to remove the respective users from our list and to replace them by other users from the same region and the same gender.

### 4.5 Dialect Annotation Task

As the dialect and the regions are known in advance to the annotators, we instructed them to double check and mark the cases in which the dialect used by a certain Twitter user appears to be from a different dialect group. This is possible despite our initial filtering based on distinctive regional keywords. We noticed that in more than 90% of the cases the profiles selected belong to the specified dialect group. For the 10% remaining, we observed many cases of people borrowing terms and expressions from other dialects.

### 4.6 Data Collection and Processing

Once we have the list of profiles ready for collection, we used Twitter Stream API and the geographic filter to make sure that the collected Tweets are within the specified region. The Twitter API restricts the maximum number of Tweets to be collected to 3240 per user. We wrote a Python script to automate the data collection effort for each of the 11 regions using the list of 100 profiles for each region.

As the data collected from social media is usually noisy despite the manual verification done by the annotators, we had to write a script to clean the collected Tweets from non-textual content such as images and URLs. Moreover, we filtered all non-Arabic content from the Data. We also discarded all retweets as well as all tweets that have less than three words.

---
[4] https://how-old.net/

## 5. Annotation Evaluation

In order to evaluate the quality of the annotation, we used the standard Inter-Annotators Agreement (IAA) measures to find out to what extent the annotators are in agreement by using the provided annotation guidelines.

The three annotators involved in the project were given a sample of 110 accounts (10 per region) to be annotated by all in a blind way, without them knowing that this particular evaluation set is also given to their colleagues.

As in similar annotation projects, the inter-annotator agreement was measured using Cohen's kappa. For this task, we believe that a value above 0.75 could be considered acceptable. At the end of the evaluation, the average Kappa values obtained by the group of the annotators were: gender annotation (0.95), age group annotation (0.80) and dialect group identification (0.92) as shown in Table 4.

| Task | Kappa Score |
| --- | --- |
| Gender Annotation | 0.95 |
| Age Annotation | 0.80 |
| Dialect Annotation | 0.92 |

Table 4: Inter-annotator agreement in terms of average Kappa score; the higher the better

As expected the age identification task is a much more difficult task, especially with the absence of clear indicators. For the dialect identification task, some annotators were confused by some similarity that exists between some dialects such as the Moroccan dialect and the Algerian dialect and also by the Qatari dialect and some other Gulf dialects.

Overall, we believe that the annotation agreement is above the acceptable range for the gender and the dialect annotation tasks and it could be improved for the age annotation task.

## 6. Conclusion

We presented Arap-Tweet, a novel and large Arabic multi-dialect Twitter corpus for the Age, gender and dialect profiling. It covers 11 regions and 16 countries in the Arab world. This corpus could be used for tasks other than the identification of age, gender and dialect. For instance, with some extra annotation, it could serve for authorship attribution, sentiment analysis, deception detection and topic detection to cite a few. In the near future, we plan to provide the Arap-Tweet corpus for the research community and we hope to receive feedback from the community on the usefulness and potential applications of that resource.

## 7. Acknowledgements

This publication was made possible by NPRP grant 9-175-1-033 from the Qatar National Research Fund (a member of Qatar Foundation). The findings achieved herein are solely the responsibility of the authors.

## 8. Bibliographical References

## 9. Language Resource References

2012, Montreal, Canada.

Zaghouani, Wajdi, Diab, Mona, Mansouri, Aous, Pradhan, Sameer and Palmer, Martha. (2010). The Revised Arabic PropBank. In proceedings of the 4th Linguistic Annotation workshop (Association of Computational Linguistics), Uppsala, Sweden.
700